# Deep Learning-Based Forecasting of Hotel KPIs: A Cross-City Analysis of Global Urban Markets

C. J. Atapattu, Xia Cui, N.R Abeynayake

Manchester Metropolitan University, Manchester, UK.

**Abstract**

This study employs Long Short-Term Memory (LSTM) networks to forecast key performance indicators (KPIs), Occupancy (OCC), Average Daily Rate (ADR), and Revenue per Available Room (RevPAR), across five major cities: Manchester, Amsterdam, Dubai, Bangkok, and Mumbai. The cities were selected for their diverse economic profiles and hospitality dynamics. Monthly data from 2018 to 2025 were used, with 80% for training and 20% for testing. Advanced time series decomposition and machine learning techniques enabled accurate forecasting and trend identification. Results show that Manchester and Mumbai exhibited the highest predictive accuracy, reflecting stable demand patterns, while Dubai and Bangkok demonstrated higher variability due to seasonal and event-driven influences. The findings validate the effectiveness of LSTM models for urban hospitality forecasting and provide a comparative framework for data-driven decision-making. The model's generalisability across global cities highlights its potential utility for tourism stakeholders and urban planners.

**Introduction**

The global urban hospitality industry has experienced dynamic transformations over the past decade, shaped by rapid urbanization, evolving traveler expectations, and external shocks such as the COVID-19 pandemic. In this highly competitive and volatile environment, accurate forecasting of hotel performance has become a critical requirement for revenue management, strategic planning, and policymaking (Song & Li, 2008; Guillet & Mohammed, 2015). Key Performance Indicators (KPIs) such as Occupancy (OCC), Average Daily Rate (ADR), and Revenue per Available Room (RevPAR) are central metrics used by hoteliers and stakeholders to assess market performance and make data-informed decisions (Chen et al., 2021; Sainaghi et al., 2018).

Traditionally, forecasting in the hospitality domain has relied on classical time series techniques, particularly Autoregressive Integrated Moving Average (ARIMA), exponential smoothing, and seasonal decomposition (Witt & Witt, 1995; Chu, 1998). While effective in stable environments, these methods are often limited in their ability to capture the non-linear, multivariate, and seasonal dynamics inherent in modern urban hotel markets (Makridakis et al., 2018). Recent advances in computational power and data availability have led to a paradigm shift toward machine learning (ML) and deep learning approaches for hospitality forecasting (Law et al., 2020; Antonio et al., 2021).

Among these techniques, Long Short-Term Memory (LSTM) networks, a form of Recurrent Neural Network (RNN), have shown superior performance in time series forecasting by effectively handling long-term dependencies and non-linear trends (Hochreiter &

Schmidhuber, 1997). LSTM has been successfully applied in tourism and hospitality forecasting to improve predictive accuracy and uncover complex temporal patterns (Zhang et al., 2021; Guo et al., 2019). For instance, a study published in the *International Journal of Hospitality Management* (2021) reported forecast improvements of up to 30% when using LSTM compared to classical models. However, most existing studies tend to focus on single destination forecasting or lack comparative analysis across diverse urban contexts.

This study addresses that gap by conducting a cross-city analysis of hotel performance forecasting using LSTM models in five global urban markets: Manchester, Amsterdam, Dubai, Bangkok, and Mumbai. These cities were selected to reflect geographic diversity and varying tourism dynamics from established Western destinations to rapidly growing Asian cities. Using monthly data from 2018 to January 2025, the study forecasts three KPIs (OCC, ADR, and RevPAR) and evaluates model performance using metrics such as Mean Absolute Percentage Error (MAPE) and Mean Squared Error (MSE), following best practices in predictive analytics (Hyndman & Koehler, 2006; Lewis, 1982).

The study contributes to both theory and practice in three important ways: (1) by demonstrating the effectiveness of LSTM models in capturing the complexities of urban hotel markets, (2) by providing a comparative framework to assess the stability and forecast ability of different cities, and (3) by offering insights that are directly relevant for tourism strategists, hotel managers, and urban planners navigating a post-pandemic recovery landscape.

**Methodology**

The methodology of this empirical study adopts a comprehensive, multi-stage approach to build a robust, generalisable forecasting model for analysing urban hotel performance using advanced machine learning techniques. Five cities, Amsterdam, Dubai, Bangkok, Manchester, and Mumbai, were selected for comparative analysis based on their global prominence as tourism and business destinations, as well as their diverse geographic, economic, and cultural contexts. Manchester represents a mature European city with steady hospitality growth, while Amsterdam is a high-demand Western European tourist hub. Dubai offers a luxury hospitality market with significant fluctuations tied to global tourism trends, Bangkok represents a Southeast Asian city with high volume and seasonality, and Mumbai provides insights into a large, price-sensitive emerging market. This diverse selection ensures that the forecasting model is tested across various hospitality dynamics and urban structures, increasing the study's generalisability.

The primary data spans from 2018 to 2025 (monthly data) and includes core hotel performance indicators from STR Global: Occupancy Rate (%), Revenue per Available Room (RevPAR), and Average Daily Rate (ADR).

Data preprocessing involved several critical steps to ensure analytical rigour. Missing values and outliers were addressed through imputation and statistical filtering. Normalisation techniques, such as min-max scaling and z-score standardisation, were applied to bring all variables onto a comparable scale. The data were then structured into monthly time series formats.

The forecasting model is based on Long Short-Term Memory (LSTM) networks, a specialised type of Recurrent Neural Network (RNN) designed to handle sequential data with long-range dependencies. Traditional RNNs struggle with vanishing gradients when learning patterns over long sequences. In contrast, LSTMs incorporate a memory cell structure controlled by input, output, and forget gates, allowing them to retain or discard information over time effectively. This makes LSTM particularly suitable for time series forecasting in complex environments such as the hotel industry, where temporal dependencies, seasonality, and abrupt changes coexist. Hyperparameters for the LSTM models were fine-tuned using a combination of grid search and Bayesian optimisation to enhance performance.

Two forecasting strategies were implemented: rolling forecasting to assess the model's adaptability over time and multi-step forecasting for projecting future KPI values at 3-, 6-, and 12-month intervals. Evaluation of model performance was conducted using widely accepted metrics, including Mean Absolute Error (MAE) and Mean Absolute Percentage Error (MAPE).

The forecasting model was developed using 80% of the data (from 2018 to late 2023) for training and the remaining 20% (late -2023 to 2025) as the testing set, enabling robust out-of-sample evaluation. To ensure the generalisability of the proposed framework, forecasting accuracy was compared across five diverse cities, Amsterdam, Dubai, Bangkok, Manchester, and Mumbai, each representing a distinct economic and tourism profile. This cross-city validation confirmed the model's adaptability and potential for wider application in global urban hospitality analytics.

## Results and Discussion

### Trends and Pandemic Recovery Analysis for KPIs

The following subsections present a detailed analysis of the five key performance indicators (KPIs) used to evaluate urban hotel performance across the five cities. These KPIs, Occupancy Rate (OCC), Average Daily Rate (ADR), and Revenue per Available Room (RevPAR), are examined in terms of their temporal patterns, the impacts of the COVID-19 pandemic, and their subsequent recovery paths. By comparing these indicators across cities, we identify both common trends and location-specific variations that highlight the diverse nature of global urban hospitality markets.

### Occupancy Rate (OCC)

The occupancy trends across Amsterdam, Bangkok, Dubai, Manchester, and Mumbai reveal distinct pandemic responses and recovery patterns shaped by their economic drivers (Figure 1). Pre-pandemic, Dubai and Amsterdam led with 75-85% occupancy rates, buoyed by robust tourism and business travel, while Amsterdam maintained stable 80-85% levels as a European hub. Mumbai showed moderate volatility (75-80%) as India's financial capital, and Manchester remained steady at 75-80% on domestic demand. The pandemic collapse hit Bangkok and Dubai hardest (15-20% in 2020) due to their tourism dependence, but Dubai rebounded fastest, surpassing pre-pandemic levels by 2023 through luxury tourism and events like Expo 2020. Mumbai recovered strongly to 75%+ by 2023, benefiting from returning business travel, and Manchester reached full recovery by mid-2022 (>75%) through domestic activity. Amsterdam lagged (70-75% through 2023) with slower business travel returns, while Bangkok struggled to exceed 75% even by 2024. The recovery highlighted Dubai and Mumbai's resilience due to diversified demand, Manchester's domestic stability, and the prolonged challenges faced by tourism-centric markets like Bangkok. Looking ahead, Dubai and Mumbai are poised to maintain 75-85% occupancy, Manchester 75-80%, and Amsterdam 75-80%, while Bangkok's recovery remains contingent on the return of international tourists, particularly from China. These trends underscore how economic diversification and policy responses shaped post-pandemic occupancy recoveries across global cities.

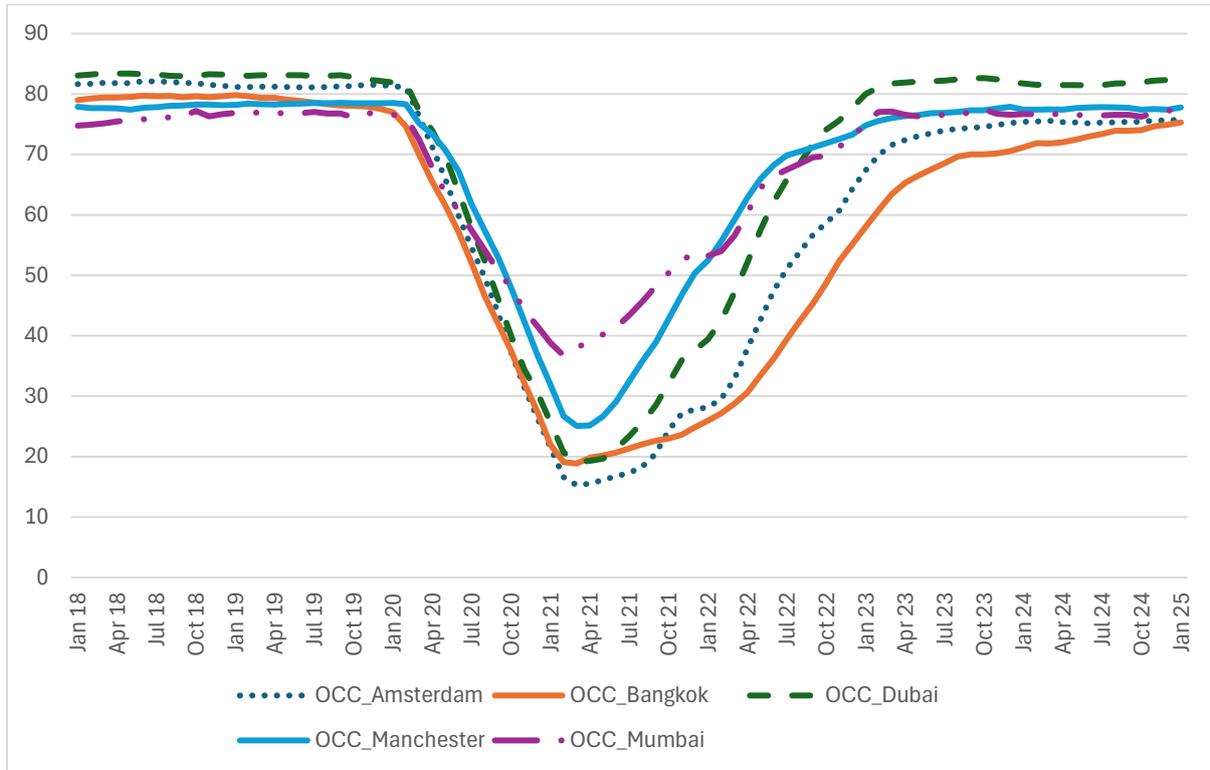

Figure 1: Monthly Occupancy Rate (OCC); Trends in Five Global Cities (January 2018 – January 2025)

**Average Daily Rate**

The Average Daily Rate (ADR) trends for these five global cities reveal distinct market behaviours and recovery patterns from January 2018 through January 2025 (Figure 2). Amsterdam began as the highest-priced market (peaking at $177 in late 2018) but suffered the most severe pandemic shock, with rates plunging 45% to $99 by March 2021. Its recovery remained incomplete, still 5% below pre-pandemic levels by 2025. Dubai showed similar premium positioning but demonstrated greater resilience - after a 33% drop to $102 in early 2021, it not only fully recovered but surpassed pre-COVID rates by mid-2022, reaching record highs above $195 by late 2023. Bangkok's trajectory was most dramatic - starting as the lowest-priced market ($98 in 2018), it collapsed completely during the pandemic ($66 by September 2021) before mounting an impressive 84% recovery to $121 by 2025. Manchester maintained remarkable stability throughout, with the shallowest decline (just 26%) and steady recovery to $115 by 2025. Mumbai emerged as the standout performer, starting mid-range at $122, it weathered the crisis better than peers (bottoming at $57) before embarking on a sustained growth trajectory that saw it overtake all cities except Dubai and Amsterdam, reflecting India's economic ascent. These patterns highlight how premium markets (Amsterdam) proved most vulnerable to global shocks, while diversified hubs (Dubai) and emerging markets (Mumbai) showed greater adaptability, and domestic-focused cities (Manchester) provided stability. The

post-2022 period particularly reveals the reordering of global hospitality markets, with traditional hierarchies being reshaped by varying recovery speeds and new economic realities.

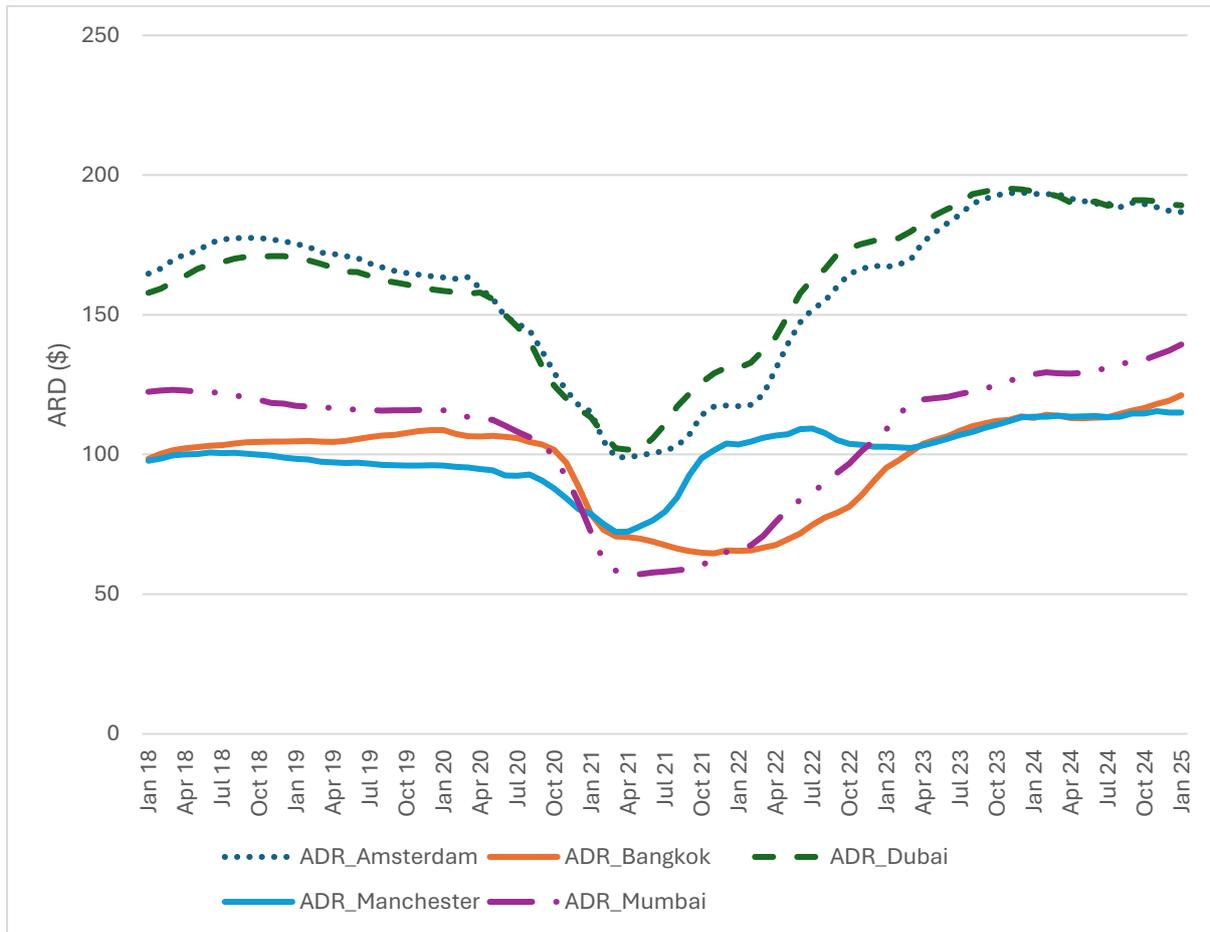

Figure 2: Monthly Average Daily Rate (ADR); Trends in Five Global Cities (January 2018 – January 2025)

**Revenue per Available Room**

The Revenue per Available Room (RevPAR) trends for these five cities from January 2018 to January 2025 reveal distinct market dynamics and pandemic recovery patterns (Figure 3). Amsterdam began as the RevPAR leader ($134 in Jan 2018) but suffered the most severe collapse during the pandemic, plummeting 82% to just $24.80 by January 2021 - the steepest decline among all cities. While it recovered to $141 by 2025, this still represented only 95% of its pre-pandemic level, highlighting lasting challenges in regaining its premium positioning.

Dubai demonstrated remarkable resilience despite its tourism dependence. After falling 79% from $131 to $29.14 during the crisis, it not only achieved full recovery by late 2022 but surpassed pre-pandemic levels, reaching record highs of $161 in October 2023, a testament to its successful reopening strategy and luxury market appeal. Bangkok's trajectory was most dramatic, starting at $77.78, it collapsed completely to $13.32 in March 2021 before mounting a strong recovery to $91.22 by 2025 (117% of pre-COVID levels), benefiting from pent-up tourism demand.

Manchester showed the steadiest performance throughout the period. With the shallowest decline (67% from $76.16 to $25.11) and consistent recovery, it exceeded pre-pandemic

RevPAR by 2023, reaching $89.45 in January 2025, outperforming Amsterdam despite starting 43% lower. This reflects the stability of its domestically driven market. Mumbai emerged as the growth champion, though starting mid-range at $91.47, it weathered the crisis better than peers (bottoming at $22.15) before surging to $107.93 by 2025 (118% of pre-COVID levels), underscoring India's economic momentum.

The post-2021 period reveals striking market reordering: while traditional leaders (Amsterdam) struggled, emerging markets (Mumbai) and adaptable hubs (Dubai) gained ground. By 2025, the RevPAR hierarchy had fundamentally shifted, Dubai led at $155.92, followed closely by Mumbai ($107.93), with Amsterdam ($141.42) barely ahead of Bangkok ($91.22) despite their vastly different starting points. These trends highlight how crisis responses, market diversification, and economic fundamentals reshaped global hospitality competitiveness through this turbulent period.

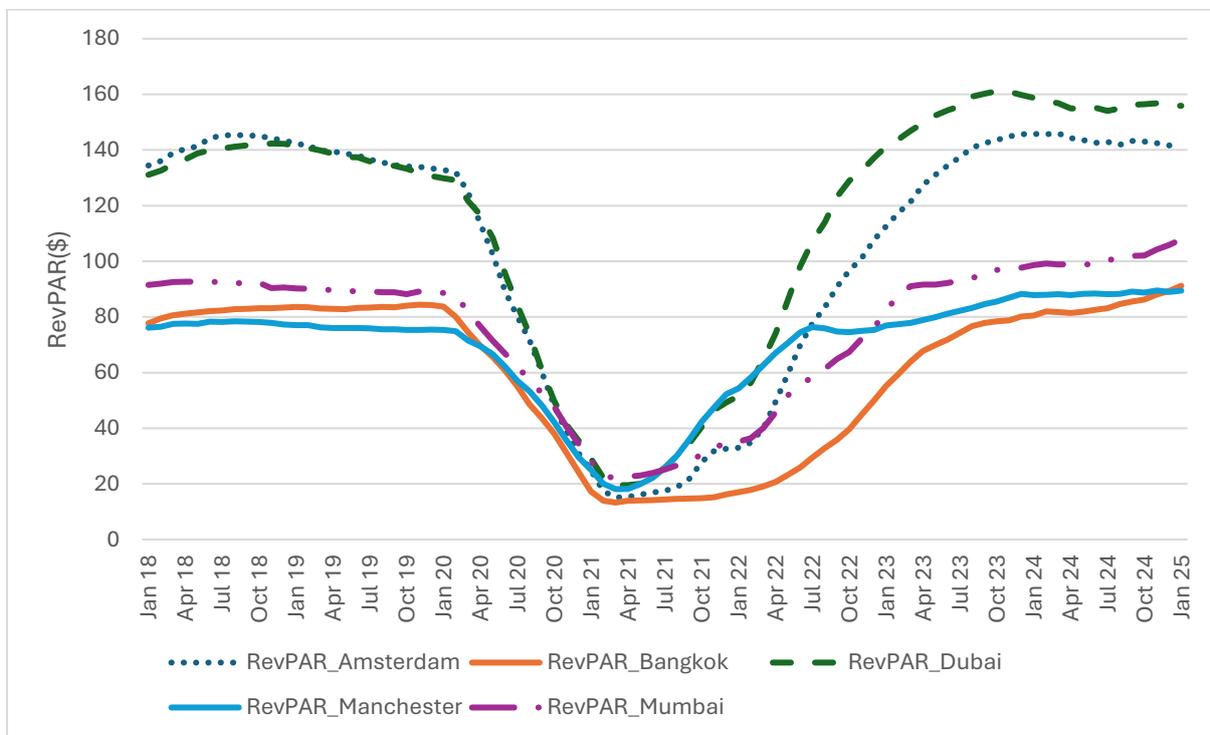

Figure 3: Monthly Revenue per Available Room (RevPAR); Trends in Five Global Cities (January 2018 – January 2025)

**Performance Evaluation of LSTM Forecasting Models**

The forecasting performance of the LSTM model was evaluated using two metrics: Mean Squared Error (MSE) and Mean Absolute Percentage Error (MAPE), for three key performance indicators (KPIs): Occupancy (OCC), Average Daily Rate (ADR), and Revenue per Available Room (RevPAR) across five cities. The results are summarised in the Table 1:

Table 1: Forecasting Accuracy of LSTM Models for Hotel KPIs Across Five Cities (Measured by MSE and MAPE)

| City | OCC | | ADR | | RevPAR | |
|---|---|---|---|---|---|---|
| | MSE | MAPE | MSE | MAPE | MSE | MAPE |
| Manchester | 6.22 | 3.10% | 40.64 | 2.80% | 47.68 | 4.46% |
| Amsterdam | 6.09 | 2.83% | 70.91 | 5.88% | 15.81 | 4.26% |
| Bangkok | 9.75 | 3.62% | 48.75 | 3.41% | 96.48 | 5.11% |
| Dubai | 4.80 | 2.75% | 14.30 | 3.06% | 8.62 | 2.72% |
| Mumbai | 7.58 | 3.49% | 3.45 | 1.08% | 23.12 | 3.93% |

The categorization of forecast predictability based on MAPE is widely used in forecasting literature to assess the accuracy and reliability of prediction models. The most frequently cited categorisation was proposed by Lewis (1982) and referenced in many forecasting studies (e.g., Hyndman & Koehler, 2006; Makridakis et al., 2018).

| MAPE (%) | Forecast Accuracy Category |
|---|---|
| < 10% | Highly accurate forecasting |
| 10% – 20% | Good forecasting |
| 20% – 50% | Reasonable forecasting |
| > 50% | Inaccurate forecasting |

According to the categorisation outlined above, the models developed using LSTM exhibit strong predictive performance.

Figures 4 illustrates the original, fitted, and forecasted patterns of the key performance indicators (KPIs) for Manchester.

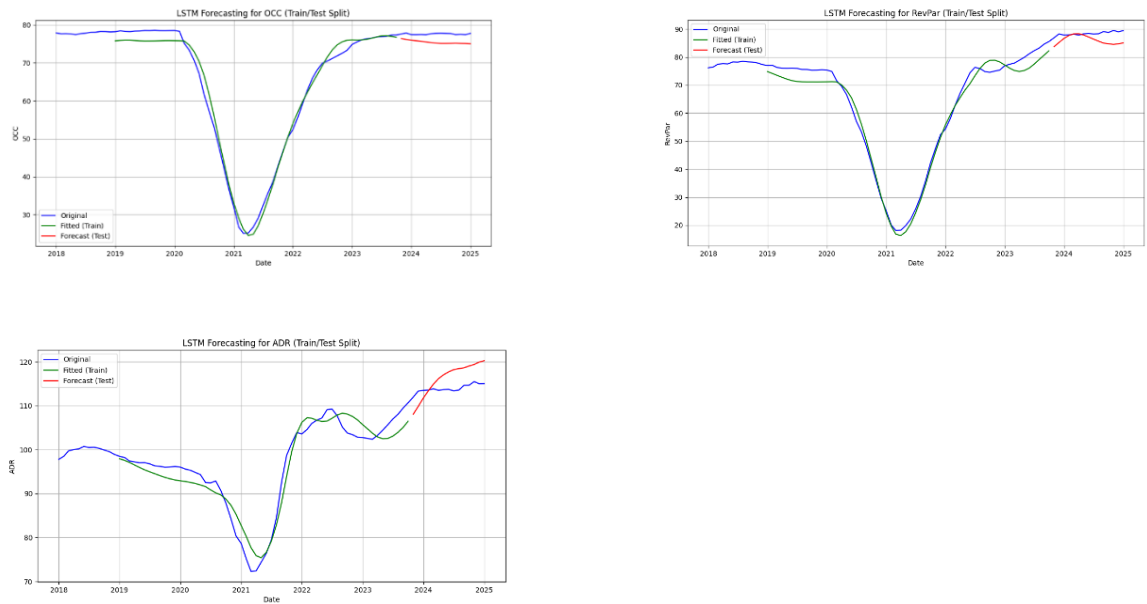

Figure 4: Manchester - LSTM Forecasting for OCC, ADR, and RevPAR

Manchester consistently exhibits low MSE and MAPE across all KPIs, particularly for RevPAR (MSE = 8.62, MAPE = 2.72%) and OCC (MSE = 4.80, MAPE = 2.75%), indicating that the LSTM model forecasts hotel performance in Manchester with high accuracy. This suggests strong temporal patterns and potentially less volatility in the city's hospitality data compared to others.

Figures 5 illustrates the original, fitted, and forecasted patterns of the key performance indicators (KPIs) for Amsterdam.

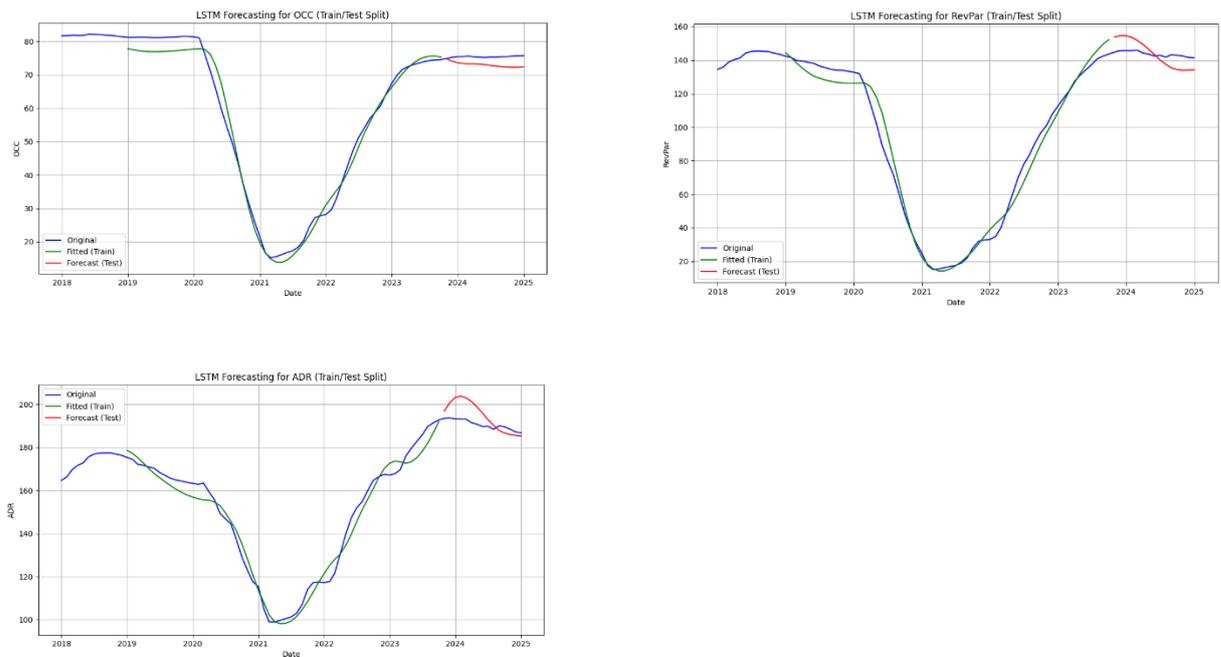

Figure 5: Amsterdam - LSTM Forecasting for OCC, ADR, RevPAR

Amsterdam performs moderately well across all three indicators, although RevPar shows relatively higher error, indicating potential room for model improvement or additional influencing factors not captured in the dataset.

Figures 6 illustrates the original, fitted, and forecasted patterns of the key performance indicators (KPIs) for Bangkok.

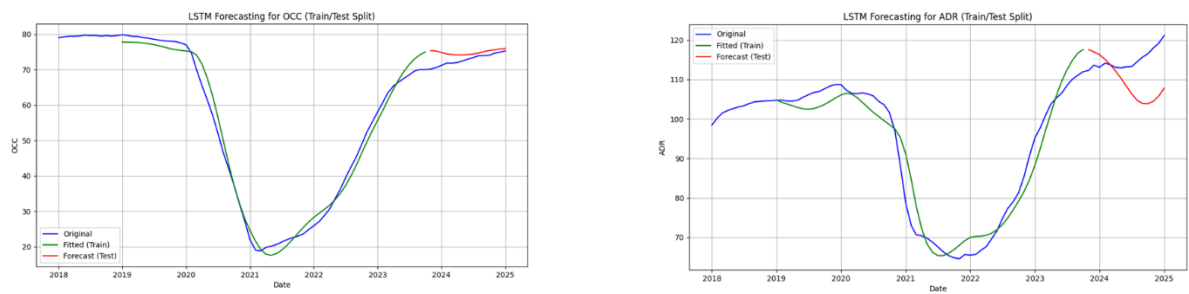

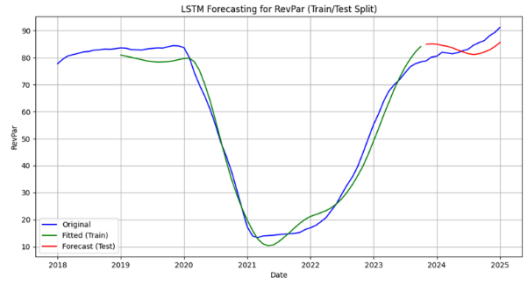

Figure 6: Bangkok - LSTM Forecasting for OCC, ADR, RevPAR

Bangkok's ADR forecasting presents challenges, with the highest MAPE (5.88%) and highest MSE (70.91) among the cities, which may reflect variability in pricing strategies or external economic influences.

Figures 7 illustrates the original, fitted, and forecasted patterns of the key performance indicators (KPIs) for Mumbai.

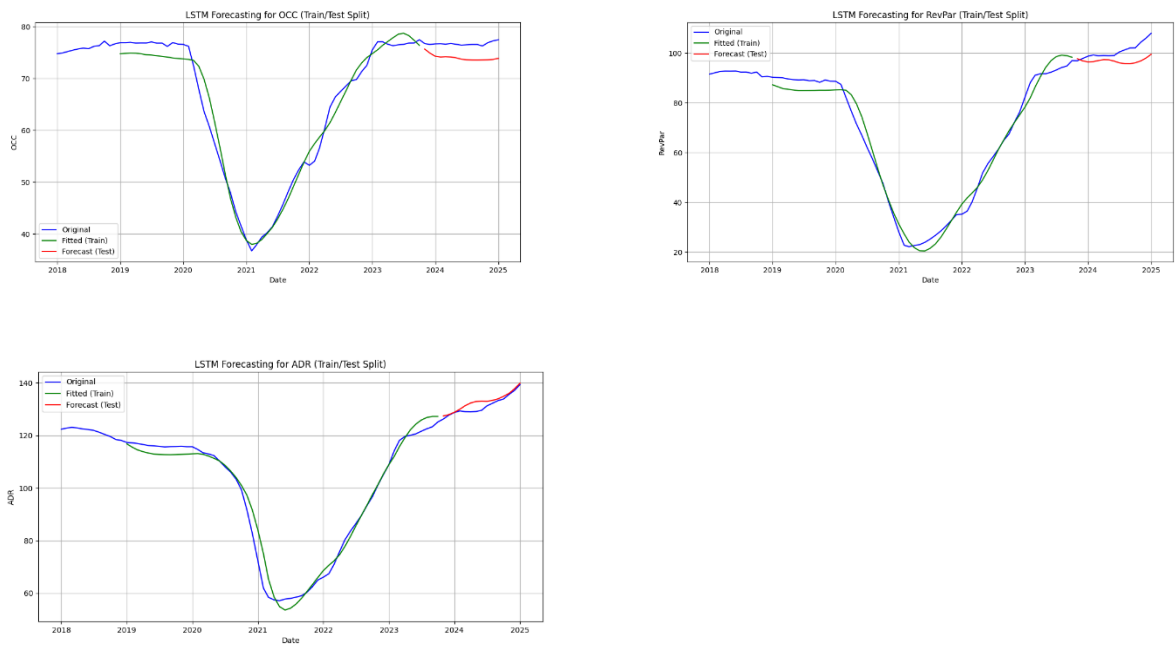

Figure 7: Mumbai - LSTM Forecasting for OCC, ADR, RevPAR

Mumbai demonstrates the lowest MAPE for ADR (1.08%), even though its MSE is also the lowest among all cities, indicating that price trends are relatively stable and predictable in Mumbai's hotel sector.

Figures 8 illustrates the original, fitted, and forecasted patterns of the key performance indicators (KPIs) for Dubai.

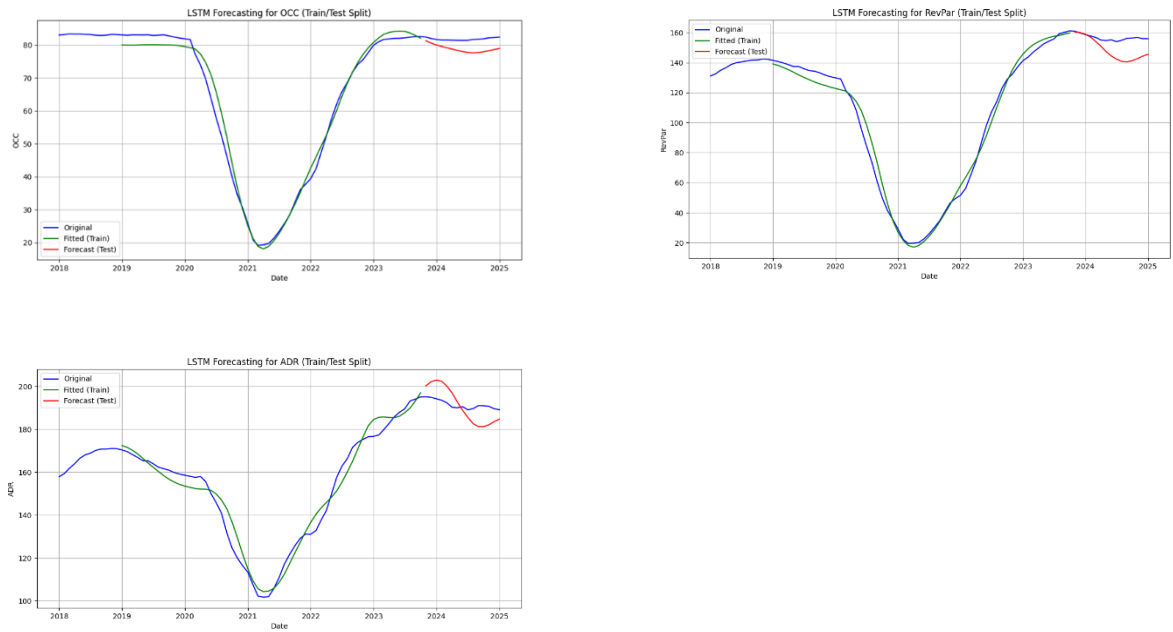

Figure 8: Dubai - LSTM Forecasting for OCC, ADR, RevPAR

Dubai shows the highest MSE and MAPE for RevPAR (MSE = 96.48, MAPE = 5.11%), implying greater complexity or unpredictability in the revenue patterns with compare to other cities, possibly due to fluctuations driven by seasonality, high-end tourism, and event-based demand.

**Conclusion**

This study demonstrates the effectiveness of advanced machine learning techniques, particularly Long Short-Term Memory (LSTM) networks, in forecasting hotel performance across diverse urban settings. By analysing three key performance indicators, Occupancy, Average Daily Rate (ADR), and Revenue per Available Room (RevPAR), for five globally significant cities (Manchester, Amsterdam, Dubai, Bangkok, and Mumbai), the research reveals both the strengths and challenges of forecasting in varying market contexts.

Manchester and Mumbai exhibited high forecast accuracy, reflecting more stable and predictable hospitality trends, while Dubai and Bangkok showed greater volatility, driven by seasonal tourism, luxury market dynamics, and international travel dependencies. Amsterdam, though traditionally a strong performer, displayed lingering post-pandemic effects, particularly in RevPAR recovery.

The study confirms that LSTM models, supported by robust time series decomposition and well-structured training/testing strategies, offer a reliable and scalable approach for urban hospitality forecasting. These findings can guide hoteliers, policymakers, and tourism strategists in enhancing operational planning, investment decisions, and post-crisis recovery efforts. Moreover, the model's adaptability across different urban profiles underscores its relevance for global application in data-driven hospitality management.